%% file: main.tex
\documentclass[runningheads]{llncs}

\usepackage{eccv}

\usepackage{eccvabbrv}

\usepackage{wrapfig}
\usepackage{tikz}
\usepackage{xcolor}

\definecolor{nearblue}{HTML}{1F77B4}
\definecolor{midorange}{HTML}{FF7F0E}
\definecolor{fargreen}{HTML}{2CA02C}
\DeclareRobustCommand{\legsquare}[1]{%
  \begin{tikzpicture}[baseline=(char.base)]
    \node[fill=#1, inner sep=1.2pt, rectangle, rounded corners=0.5pt] (char) {\vphantom{A}};
  \end{tikzpicture}%
}

\usepackage[accsupp]{axessibility}  % Improves PDF readability for those with disabilities.

\usepackage{hyperref}

\usepackage{orcidlink}
\usepackage{booktabs}
\usepackage{colortbl} % For the column highlighting

\begin{document}

\title{Comprehensive Robustness Analysis of LiDAR-based 3D Object Detection in Autonomous Driving} 

\titlerunning{Robustness Analysis of LiDAR-based 3D-OD in Autonomous Driving}

\author{Adwait Chandorkar\inst{1}\orcidlink{0000-0001-9805-5912} \and
Kai Krink\inst{1}\protect\textsuperscript{*}\orcidlink{0009-0005-1637-6859} \and
Yerdana Maulenbay\inst{1,2}\protect\textsuperscript{*,\,\protect\(\dagger\protect\)}\orcidlink{0009-0000-2367-5447} \and
Hasan Tercan\inst{1}\orcidlink{0000-0003-0080-6285} \and
Tobias Meisen\inst{1}\orcidlink{0000-0002-1969-559X}
}
\authorrunning{A. Chandorkar et al.}
\institute{Institute for TMDT, University of Wuppertal, Germany \\
\email{\{chandorkar, kai.krink, tercan, meisen\}@uni-wuppertal.de}\\
%\url{https://tmdt.uni-wuppertal.de/en/} \and
\and
IKB Faculty of Science, University of British Columbia, Canada\\
\email{yerdana@student.ubc.ca}}

\maketitle
\begingroup
\renewcommand{\thefootnote}{\fnsymbol{footnote}}
\footnotetext[1]{Equal contribution}
\footnotetext[4]{The author contributed to this work as part of the DAAD RISE program.}
\endgroup
\input{sec/0_abstract}
\input{sec/1_introduction}
\input{sec/2_related_work}
\input{sec/3_methodology}
\input{sec/4_results}
\input{sec/5_conclusion}

\bibliographystyle{splncs04}
\bibliography{main}
\end{document}

%% file: sec/0_abstract.tex
\begin{abstract}
Recent advancements in LiDAR-only 3D object detection have demonstrated improved detection accuracy over benchmark datasets. However, the adversarial robustness of these models remains untested. Very few adversarial robustness studies exist for LiDAR-only 3D object detection and unfortunately, even they are limited to legacy models. Moreover, there is a systemic gap in the existing evaluation frameworks that rely simply on mAP ignoring other structural and predictive factors. To fill this gap, we propose a holistic framework that evaluates adversarial robustness using two structural factors (point cloud density and point cloud localization) and three predictive factors (misclassification, localization error, distance from ego). Using this framework, we perform an empirical study and critical analysis on recent and legacy state-of-the-art models using adversarial attacks specifically designed for LiDAR-based models. Our key finding is that high-capacity, voxel-based detectors are more susceptible to structured coordinate perturbations than pillar-based detectors. Additionally, non-anchor-based detectors demonstrate poor adversarial robustness, which necessitates rethinking model training techniques. Overall, our results demonstrate that recent models are as vulnerable to adversarial attacks as their predecessors. Therefore, we argue that there is a need to improve the evaluation benchmarks for 3D object detection that not only reward architectural modifications for improving detection accuracy, but also evaluate whether the design choices improve adversarial robustness. 
\keywords{Adversarial attacks \and 3D object detection \and Robustness analysis}
\end{abstract}

%% file: sec/1_introduction.tex
\section{Introduction}
\label{sec:intro}

Perception has long been one of the most critical tasks in autonomous driving systems. Over the years, we have seen significant progress in perception systems, both in standalone sensor systems (\eg LiDAR-only or camera-only) and in multi-sensor systems. That said, LiDAR-based systems remain the preferred choice, thanks to the robustness of the sensor and its ability to efficiently capture depth information \cite{robustness_lidar_1,robustness_lidar_2, robustness_lidar_3}. From the perspective of representation learning, LiDAR-only object detection (LiDAR-OD) models can be broadly classified into point-based methods \cite{PointRCNN, 3DSSD}, grid-based methods \cite{pointpillars, pillarnet,voxelnet,voxelnext,second,pillarnext,pillarnest,focalformer3d,Chandorkar}, and hybrid methods \cite{pvrcnn,pvrcnn++,parta2,uni3detr} based on how the raw point clouds are processed. Grid-based methods, mainly, have outperformed other methods on demanding benchmark datasets such as KITTI \cite{kitti}, nuScenes \cite{nuscenes} and Waymo \cite{waymo}. These methods have not only demonstrated higher detection accuracy but also achieved higher inference speeds demonstrating significant progress toward the realization of fully autonomous driving in real-world environments.

Nevertheless, as previous works \cite{robo3d,vul_analy_lidar_3d_od, robustness_lidar_1} have reported, model improvements often deliver strong results on clean datasets, but their robustness remains significantly compromised when subjected to corrupt or out-of-distribution (OoD) examples. One way of evaluating this is to test the robustness of these models against adversarial attacks. Robustness analysis using adversarial attacks was initially rooted in the 2D image domain \cite{goodfellow2014explaining, towards_dl_adv}. However, recently there has been increased research in the 3D domain \cite{generating_adv_pc,imperceptible_adv_att_3d_od,Und_robust_3d_od_bev,lidattack,iou_s,physically_ad_lidar_od}. Although most of these works have targeted the exploitation of vulnerabilities in existing 3D-OD and measured the resulting performance degradation, the current literature fails to investigate whether recent architectures \cite{pillarnext, pillarnet, pillarnest, focalformer3d} have actually improved robustness or if they inherit similar structural vulnerabilities from their predecessors. In addition, we identify three more gaps in existing studies: \emph{first}, an over-reliance on benchmarking legacy 3D-OD models; \emph{second}, an evaluation scope restricted almost exclusively to Average Precision (AP), neglecting critical localization errors (\eg, translation, scale) and factors like point cloud density or proximity to the ego vehicle; and \emph{third}, a dependence on classical adversarial attacks such as FGSM \cite{goodfellow2014explaining} and PGD \cite{pgd} rather than attacks designed for LiDAR-based 3D-OD.

In this work, we address these shortcomings by presenting a comprehensive benchmark suite to evaluate the adversarial robustness of 3D-OD across commonly benchmarked datasets such as nuScenes and Waymo. We hypothesize that, first, \emph{SoTA 3D-OD architectures exhibit a fundamental lack of resilience against coordinated PC perturbations} and second, \emph{a holistic evaluation requires decoupling structural robustness (geometric precision) from predictive robustness (detection confidence)}. To our knowledge, this represents the first exhaustive study of its kind for LiDAR-based 3D-OD. We evaluate different white-box and black-box attacks, designed specifically for LiDAR-based 3D-OD, to expose inherent model weaknesses. Moving beyond standard metrics like Attack Success Rate (ASR) or mAP degradation, we propose five distinct benchmarks to quantify the impact across point cloud (PC) density, PC localization, object classification, object localization, and the object's proximity from the ego vehicle. Our study seeks to establish whether the architectural improvements in latest SoTA 3D-OD yield improved adversarial robustness or the models remain vulnerable like previous architectures. Our main contributions are summarized as follows:
\begin{itemize}
\item We propose a comprehensive adversarial robustness analysis of LiDAR-based 3D-OD models by benchmarking our results on two \emph{latest}  3D-OD and comparing their results with legacy  models.
\item Our study analyzes both the structural robustness, the impact of PC modifications, and the predictive robustness of the baselines and addresses the gaps that arise from architectural design choices.
\item We identify the main reason for this adversarial susceptibility to \emph{Representation Fragility} in the existing datasets: the models prioritize the memorization of specific geometric templates and canonical point-cloud densities prevalent within the training distribution, rather than the learning geometric inconsistencies. This approach yields superior performance on clean datasets but collapses under structured noise.
\end{itemize}

%% file: sec/2_related_work.tex
\section{Related Work}
\label{sec:related_work}

\subsection{Advancements in grid-based 3D object detection}
  
The grid-based architecture comprises an \emph{encoder} that transforms sparse point clouds into either voxels or pillar-based pseudo-images, followed by a \emph{backbone} for multi-level feature extraction, a \emph{neck} for feature fusion, and a \emph{head} for regressing 3D bounding boxes and class probabilities. Voxel-based methods \cite{voxelnet, second, voxelnext} partition the point clouds into 3D voxels using either 3D CNNs as proposed by Zhou \etal in VoxelNet \cite{voxelnet} or 3D SpConv proposed by Yan \etal in SECOND \cite{second}. A recent voxel-to-object method VoxelNext \cite{voxelnext}, introduced by Chen \etal, proposes direct end-to-end 3D-OD from voxel features, bypassing the need for post-processing. Despite advancements in reducing computational demands, voxel-based detectors still entail high computational costs and lower inference speeds. On the other hand, pillar-based methods provide a more efficient balance between inference speed and detection accuracy. Building on PointPillars \cite{pointpillars}, PillarNext \cite{pillarnext} and PillarNest \cite{pillarnest} replace pure max pooling with combined max–average pooling to mitigate feature loss. They further substitute conventional CNN backbones with sparse convolutions (SpConv), improving computational efficiency \cite{pillarnet, pillarnext}. To bypass the deployment challenges of SpConvs — specifically irregular memory access and quantization — PillarNest \cite{pillarnest} utilizes a ConvNext \cite{convnext} backbone pre-trained on images and scaled for point clouds, while FastPillars \cite{fastpillars} implements a re-parameterized ResNet-34 for efficient scaling. Although the \emph{encoder} varies significantly for voxel and pillar-based methods, the \emph{head} is mostly similar. Initially, grid-based methods \cite{pointpillars, voxelnet, second} relied on computationally heavy anchor-based detectors such as RPN or SSD. Subsequently, CenterPoint \cite{centerpoint} proposed an anchor-free design, \emph{CenterHead}, which detects objects through center point estimation. More recently, transformer-based heads (\eg, FocalFormer3D \cite{focalformer3d}, PillarTrack \cite{pillartrack}) have achieved superior performance on nuScenes \cite{nuscenes} and Waymo \cite{waymo}. In this study, we utilize 3D-OD models with enhanced \emph{encoder} and \emph{decoder} structures as baselines, with further justification provided in \cref{sec:method_3d_od}. On these baselines, we evaluate whether the enhancements have improved adversarial robustness or if it is similar to their predecessors.

\subsection{Adversarial attacks on 3d object detection}

While classical gradient-based attacks like FGSM and PGD provide a baseline for adversarial robustness, they fail to account for the sparsity, sensor physics, and non-differentiable pre-processing (\eg, voxelization and pillarization) inherent in LiDAR-based 3D-OD pipelines. Classical methods often produce \emph{point-wise noise} that is filtered out by the structural encoding in the \emph{encoder}. In contrast, Abraha \etal \cite{convnet_gen_adv_perturb} introduce a gradient-free attack that exploits detector knowledge via a ConvNet-based perturbation generator. Chen \etal \cite{iou_s} propose a white-box method optimizing targeted perturbations through IoU minimization. Chen \etal \cite{lidattack} propose a black-box genetic simulated annealing (GSA) attack that generates adversarial objects camouflaged within the scene. Furthermore, Long \etal \cite{box_attack} introduced a white-box non-end-to-end attack that manipulates feature saliency by restraining critical detection features while amplifying non-contributory features to diminish accuracy with reduced reliance on specific network architectures. We choose attacks that are specifically designed for LiDAR-based 3D-OD as our baseline and provide additional reasoning in \cref{sec:method_adv}.

\subsection{Robustness analysis of 3d object detection}

Zhang \etal \cite{comp_study_robust_lid_3d_od} provided the first systematic analysis of LiDAR-only architectures, assessing attack efficacy, transferability, and defense performance. Further, Dong \etal \cite{bench_robust_3d_od_common} benchmarked various modalities against 27 physically realizable environmental corruptions. Recently, Eskandar \cite{Emp_study_gen_lidar_3d_od} investigated domain adaptation factors that affect robustness, such as voxel encoding, resolution, and data augmentation. Despite these advancements, current adversarial robustness analysis often suffers from two critical limitations: the use of legacy adversarial attacks instead of recent methods and the lack of benchmarking on the latest SoTA 3D-OD models for evaluation. Our work overcomes these limitations and identifies the architectural choices that make modern SoTA 3D-OD vulnerable to adversarial attacks.

%% file: sec/3_methodology.tex
\section{Methodology}
\label{sec:method}

\begin{figure}[tb]
  \centering
  \includegraphics[height=4cm]{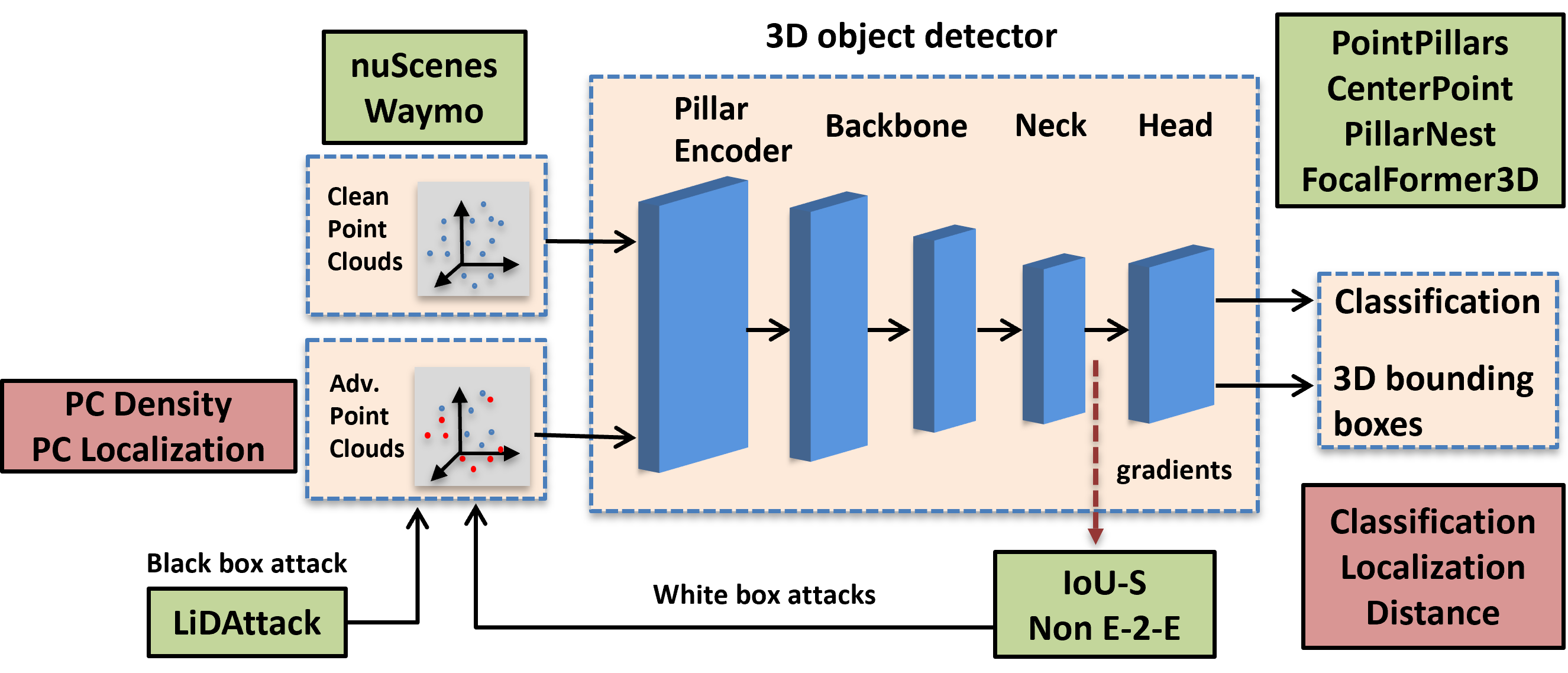}
  \caption{Methodology}
  \label{fig:3_methodology}
\end{figure}

We design our study based on three foundations: 1) we use \emph{four} baseline models (\emph{two} latest 3D-OD models and \emph{two} classical 3D-OD models) for improvement comparison, as explained in \cref{sec:method_3d_od}; 2) we use \emph{five} latest adversarial attacks that have demonstrated improved performance over classical attacks, as explained in \cref{sec:method_adv}, and 3) we propose a comprehensive benchmark involving \emph{five} criteria to study the impact on our baselines, as explained in \cref{sec:method_eval}. \cref{fig:3_methodology} demonstrates our entire setup. Our investigation is conducted on two standard benchmarks: nuScenes and Waymo. For baseline models that were originally not trained on these datasets, we train the models from scratch and report the resultant mean Average Precision (mAP) in \cref{tab:model_taxonomy} to ensure performance parity across all baselines. These initial metrics establish an upper-bound performance reference under clean point cloud conditions. We subsequently expose all baselines to four white-box and one black-box adversarial attack to generate perturbed inputs. Model robustness is then evaluated along two complementary dimensions: point cloud (PC) density and PC localization capture \emph{structural robustness}, whereas classification, bounding-box localization, and distance-based metrics quantify degradation in \emph{predictive performance}.

\begin{table}[tb] 
    \caption{Taxonomy of the models and datasets used in our study. For the models where the original implementation (O) was available, we use the pre-trained weights. For others, we train (T) them from scratch.}
    \centering 
    \scriptsize
    % 1. Increase vertical spacing between rows
    \renewcommand{\arraystretch}{} 
    % 2. Stretch table to text width using tabular*
    \begin{tabular*}{\textwidth}{@{\extracolsep{\fill}} lll c c c c @{}} % Removed vertical bars for a cleaner look
        \toprule
        \textbf{Model} & \textbf{Encoder} & \textbf{Head} & \multicolumn{2}{c}{\textbf{nuScenes}} & \multicolumn{2}{c}{\textbf{Waymo}} \\
        \cmidrule(lr){4-5} \cmidrule(lr){6-7}
        & & & Impl. & mAP $\uparrow$ & Impl. & mAP $\uparrow$ \\   
        \midrule
        PointPillars \cite{pointpillars} & Pillar & RPN & T & 35.9 & T & 58.4 \\
        PillarNeSt \cite{pillarnest} & Pillar & CenterHead & O & 66.9 & T & 66.2 \\
        \midrule
        CenterPoint \cite{centerpoint} & Voxel & CenterHead & O & 58.0 & O & 67.8 \\
        FocalFormer3D \cite{focalformer3d} & Voxel & Transformer & O & 70.5 & O & 71.9 \\
        \bottomrule
    \end{tabular*}
    \label{tab:model_taxonomy}
\end{table}

\subsection{Baseline 3d object detectors}
\label{sec:method_3d_od}

Our experimental design is based on a strategic selection of baselines that serve as critical testbeds for our two hypotheses. To evaluate first hypothesis, the fundamental vulnerability of SoTA 3D-OD to coordinated PC perturbations, we incorporate PillarNeSt \cite{pillarnest} and FocalFormer3D \cite{focalformer3d}. As an evolution of the pillar-based paradigm, PillarNeSt demonstrates a strong feature extraction efficiency, utilizing a ConvNeXt backbone and a dual-path (max and average) pooling strategy to minimize information entropy in \emph{encoder}. It also outperforms other recent pillar-based OD, such as PillarNet \cite{pillarnet}, PillarNext \cite{pillarnext}, and FastPillars \cite{fastpillars} on nuScenes dataset. By selecting the leading model of this lineage, we aim to determine if enhanced \emph{encoder} architecture increases resilience to PC perturbations. FocalFormer3D represents a shift toward transformer-based detection \emph{head}, specifically designed to mitigate false negatives (FN) through a Hard Instance Probing (HIP) pipeline. FocalFormer3D has significantly outperformed other transformer-based OD such as OcTr \cite{octr}, Li3DeTr \cite{erabati2022li3detrlidarbased3d} and FusionViT \cite{fusionvit} on both nuScenes and Waymo. This selection allows us to investigate whether architectural improvements in \emph{head} improve geometric precision or merely mask inherent structural vulnerabilities. Finally, we compare the robustness of these models with CenterPoint \cite{centerpoint} and PointPillars \cite{pointpillars}. These models provide the necessary contrast between modern and legacy models. 

\subsection{Baseline adversarial attacks}
\label{sec:method_adv}

For the white-box setting, we first use the IoU-S \cite{iou_s} attack. The central idea is to target the localization capability. A 3D-OD predicts an object class along with 3D bounding box co-ordinates, i.e. localize the objects. Based on the predicted class probability and the IoU-score, a \emph{confidence} score is generated using the formula $\text{confidence} = \sqrt{P_{\text{class}} \times \text{IoU}_{\text{score}}}$. The IoU-S attack learns to generate adversarial point clouds that minimize this confidence of the attacked model. We use all the three attack paradigms proposed by IoU-S - attachment (PA), detachment (PD), and perturbation (PB). The attachment attack injects new points into the existing point cloud, the detachment attack drops points, while the perturbation attack shifts the coordinates of the raw point clouds. Then, we use the Non End-to-End (NE) \cite{box_attack} as the second white-box attack. The method overcomes the limitations of existing white-box methods: dependence on network structure and the resulting inefficient transferability of attacks. Additionally, it proposes an adversarial loss function on the feature space that restrains the features with a high contribution towards the detection task and simultaneously amplify the features with a lower contribution. Finally, we use LiDAttack (LiD) \cite{lidattack} for the black-box setting. It is a novel approach that uses genetic simulated annealing (GSA) to generate a stealthy and concealable attack by strictly constraining the distance between the perturbed points. 

An inherent challenge with adversarial attacks is that they are not architecture-agnostic and possess bias against certain design choices. To address this bias, we ensure that our attack selection remains fair and there exists at least one attack that is biased towards each of our baseline models. The PA attack affects the depth information in the point cloud (PC) and is hence particularly effective against pillar-based models. The PB attack alters the structural representation, making them effective against voxel-based models whereas the PD attack alters the density of PC representations, effectively posing higher challenges for heatmap-based detectors such as CenterPoint. The NE attack targets the intermediate feature maps rather than the PCs and thus disrupts the spatial continuity. This makes the attack especially stronger against heatmap-based detectors that rely on local maximum (\emph{peak}) of a predicted heatmap. Finally, the black-box LiD attack is a more generic attack that is not strongly biased towards any model. However, we hypothesize that the \emph{max-pooling} used in pillar \emph{encoder} makes them more vulnerable to LiD attack. The detailed formula, methodology and budget of these attacks is attached in the Appendix.

\subsection{Evaluation Criteria}
\label{sec:method_eval}

\subsubsection{Impact on Classification.} This is evaluated based on two factors: firstly, classical misclassification, and secondly, the impact on confidence. As highlighted by Wang \etal \cite{towards_stable_3d_od_confidence}, unstable confidence scores result in flickering predictions thereby impacting detection and tracking performance of 3D-OD. Hence, we assess the average drop in confidence along with average drop in mAP after subjecting the model to an adversarial attack. We also evaluate the models based on change in false positives (FP) and false negatives (FN). Our goal is to evaluate whether models are especially vulnerable to certain classes. Hence, following the works of \cite{safety_critical_1, safety_critical_2, safety_critical_3}, we first categorize Cyclists, Pedestrians and Motorcyclists as \textbf{Safety Critical} classes, since they are most involved in accidents and carry a high chance of human fatality. Then, we report the impact on \emph{Cars} and \emph{Safety Critical} classes in the \cref{sec:results_class}. 

\subsubsection{Impact on Localization.} A high IoU does not always indicate accurate localization. The bounding box could be either shifted, incorrectly scaled, or have a shifted orientation. While evaluating the performance of 3D-OD, these factors are addressed to some extent in nuScenes and Waymo. However, they are not uniform across all datasets. To address this, we calculate the \emph{translation}, \emph{scale}, and \emph{orientation} errors based on the predicted bounding boxes for all the models. To the best of our knowledge, our work is the first to factor in bounding box localization errors to evaluate the effectiveness of adversarial attacks.

\subsubsection{Impact on object distance from ego.} An inherent problem with LiDAR is that the PC of objects near the ego-vehicle are denser than that of objects which are further away. Therefore, the detection accuracy of near-ego objects is naturally higher than that of far-away objects. By assessing the impact of attacks relative to the object's distance from the ego vehicle, we can evaluate whether the models are robust enough to detect \emph{near-ego} objects.

\subsubsection{Impact of Point Cloud Density.} The adversarial perturbations generated by our baseline, especially attachment (PA) and detachment (PD) attacks, inherently alter the spatial distribution of the input, directly modifying the PC density. From a conventional perspective, 3D-OD performance is expected to improve with a high PC density and degrade under conditions of high sparsity. By systematically evaluating model efficacy against varying input densities, we investigate whether this heuristic holds under adversarial conditions or if the induced structural modifications expose vulnerabilities.

\subsubsection{Impact of Point Cloud localization.} 
\begin{wrapfigure}[12]{r}{0.28\textwidth}
  \centering
  \includegraphics[width=0.28\textwidth]{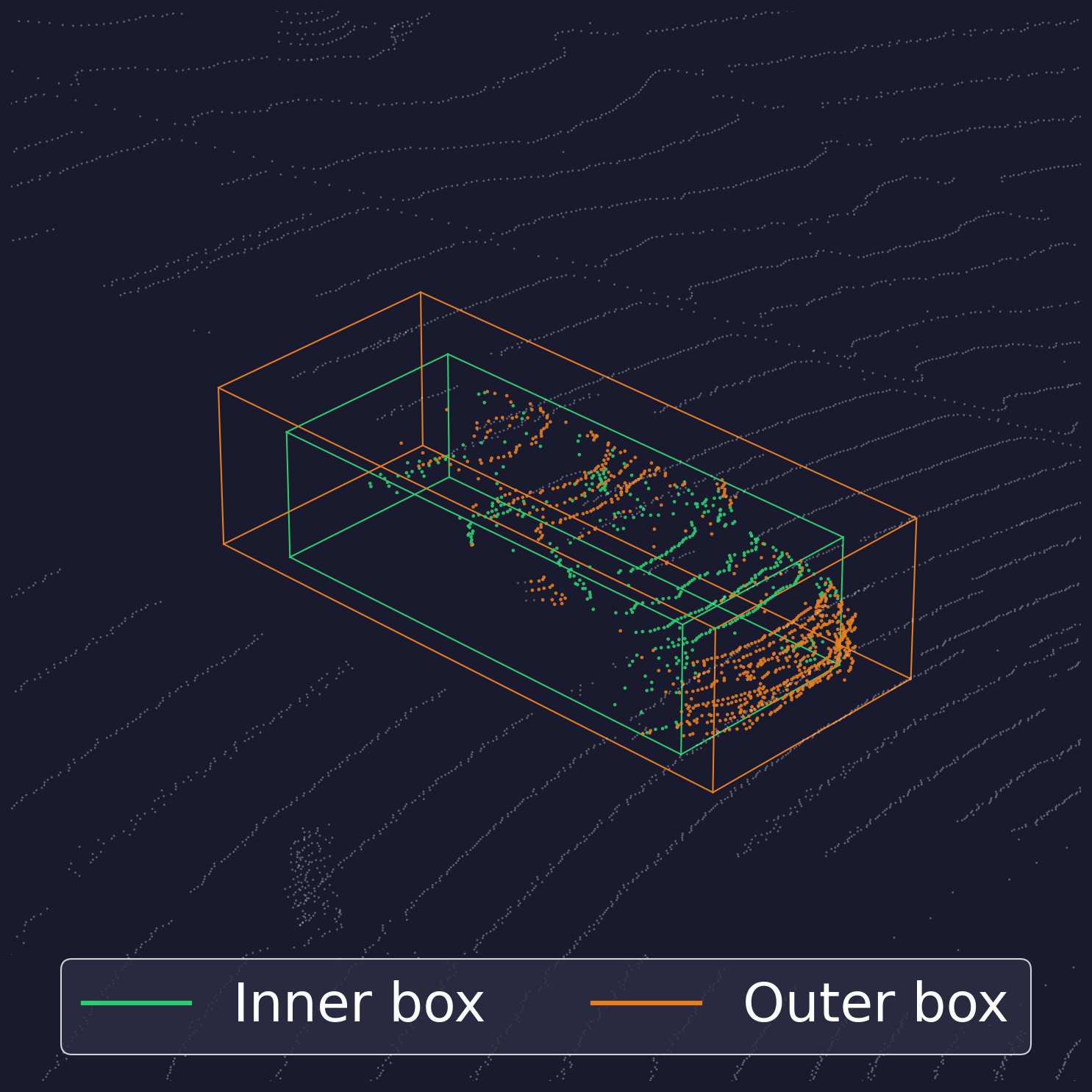}
  \caption{Visualization of inner and outer PCs for a \emph{Car} in nuScenes.}
  \label{fig:4_vis_pc_inner_outer}
\end{wrapfigure}
In order to assess the maximum impact of the modifications to the PC, we split the PCs within the ground truth box into two categories before generating the adversarial PC. As shown in \cref{fig:4_vis_pc_inner_outer}, we create a secondary 3D bounding box within the ground truth, downscaled by an empirically determined factor based on each dataset. The scaling threshold for the inner bounding boxes is set to \emph{0.8} and the empirical study to determine these thresholds is presented in \cref{sec:4_sensitivity_analysis}. We classify the points outside this scaled box as \emph{outer} points and the ones within the box as \emph{inner} points. As observed in the figure, the \emph{outer} points generally represent the object's structural contour and extremities (edges and corners), whereas the \emph{inner} points represent the volumetric density of the object. We then evaluate the impact on model performance when the attacks affect the inner and outer points. 

%% file: sec/4_results.tex
\section{Results}
\label{sec:results}

Numerous abbreviations are used throughout the following sections. To provide a single reference point, we provide a brief overview of these abbreviations here. Model abbreviations - \textbf{CP} : CenterPoint, \textbf{FF}: FocalFormer3D, \textbf{PN}: PillarNest, \textbf{PP}: PointPillars. The attack abbreviations - \textbf{PA}: Attachment, \textbf{PD}: Detachment, \textbf{PB}: Perturbation, \textbf{LiD}: LiDAttack and \textbf{NE}: Non End-2-End. We report the success of attacks using the attack success ratio (ASR). However, our method of calculating ASR is different from the method used by current approaches. We observed that existing studies report attacks as successful only if they induce misclassification but not if there is a significant drop in confidence albeit with a correct classification. Secondly, ASR calculation involves all missed detections even though they were missed by the 3D-OD on clean dataset. We argue, this results in inflated ASR and does not factor in genuine impact. To overcome these challenges, the first distinction is that we exclude the objects that were undetected by the model on the clean point cloud from the calculation. Second, we consider an attack successful if it induces misclassification or if the prediction is correct but confidence < 0.15. Otherwise, the attack is unsuccessful. In the subsequent sections, we present the most relevant results and visualizations, while additional supporting visualizations can be found in the Appendix.

\subsubsection{Experimental Setup.}
\label{sec:setup}
All models were trained and evaluated in PyTorch \cite{pytorch} using the mmdetection3d \cite{mmdet3d} codebase. Additional details associated with adversarial training, model training and configurations are available in the Appendix.
\subsection{Impact on Classification}
\label{sec:results_class}

\cref{tab:4_imp_class} lists the main results of our experiments. Our evaluation across the nuScenes and Waymo datasets reveals a bifurcated landscape of vulnerability between pillar-based models, PP and PN, and high-capacity detectors, CP and FF. Our key findings are:
\begin{table}[tb]
    \centering
    \caption{Comprehensive Adversarial Robustness Evaluation: Attack results across nuScenes and Waymo datasets. The values represent relative AP drop / relative confidence drop across different object classes and attack types. Negative values indicate improvement.}
    \label{tab:4_imp_class}
    \scriptsize
    \renewcommand{\arraystretch}{1}
    % --- Sub-table 1: nuScenes ---
    \begin{subtable}{\textwidth}
        \centering
        \caption{nuScenes}
        \label{tab:4_imp_class_nusc}
        \setlength{\tabcolsep}{3pt}
        \resizebox{\textwidth}{!}{%
        \begin{tabular}{l *{5}{cccc}}
            \toprule
            \textbf{Model} & \multicolumn{5}{c}{\textbf{Car}} & \multicolumn{5}{c}{\textbf{Safety Critical}} \\
            \cmidrule(lr){2-6} \cmidrule(lr){7-11}
            & LiD (\%) & PA (\%) & PB (\%) & PD (\%) & NE (\%) & LiD (\%) & PA (\%) & PB (\%) & PD (\%) & NE (\%) \\
            \midrule
            \textbf{CP} & -0.1/-0.3 & 3.9/6.2 & 65.5/41.8 & 15.1/2.5 & 16.6/25.7 & 0.9/-0.8 & 21.9/9.9 & 85.8/15.5 & 31.7/2.9  & 41.9/7.7 \\
            \textbf{FF} & 0.3/-2.0 & 7.8/17.7 & 38.9/59.2 & 20.6/2.5 & 17.5/37.1 & 2.6/-3.4 & 31.8/21.5 & 81.5/62.0 & 34.8/4.8 & 30.1/14.3 \\
            \textbf{PN} & 0.0/-0.2 & 32.2/4.6 & 20.2/6.4 & 24.6/1.3 & 8.0/12.6 & 0.4/-0.4 & 54.8/6.4 & 38.9/10.1 & 41.1/1.1 & 19.4/3.7 \\
            \textbf{PP} & 0.0/-0.1 & 85.4/42.2 & 10.0/18.7 & 25.9/5.9 & 8.6/19.3 & 1.0/-1.0 & 36.5/9.2 & 23.9/8.4 & 17.7/0.3 & 26.1/4.1 \\
            \bottomrule
        \end{tabular}}        
    \end{subtable}
    
    % --- Sub-table 2: Waymo ---
    \begin{subtable}{\textwidth}
        \centering
        \caption{Waymo}
        \label{tab:4_imp_class_waymo}
        \setlength{\tabcolsep}{3pt}
        \resizebox{\textwidth}{!}{%
        \begin{tabular}{l *{5}{cccc}}
            \toprule    
            \textbf{Model} & \multicolumn{5}{c}{\textbf{Car}} & \multicolumn{5}{c}{\textbf{Safety Critical}} \\
            \cmidrule(lr){2-6} \cmidrule(lr){7-11}
            & LiD (\%) & PA (\%) & PB (\%) & PD (\%) & NE (\%) & LiD (\%) & PA (\%) & PB (\%) & PD (\%) & NE (\%) \\
            \midrule
            \textbf{CP} & 0.0/-0.0 & 0.0/14.8 & 91.0/25.6 & 47.8/21.3 & 82.0/21.6 & 0.0/-0.1 & 25.3/17.6 & 96.3/28.9 & 71.3/13.8 & 81.0/20.1 \\
            \textbf{FF} & 0.2/-0.0 & 8.0/4.5 & 88.5/11.5 & 7.8/0.3 & 82.1/9.7 & 0.1/-0.2 & 19.4/19.2 & 61.1/12.5 & 24.5/7.6 & 56.7/8.3 \\
            \textbf{PN} & 1.6/-0.3 & 24.0/6.2 & 18.6/7.3 & 2.3/-0.7 & 12.8/5.4 & 1.3/-0.3 & 24.5/19.7 & 13.8/24.7 & 18.8/6.1 & 11.1/21.3 \\
            \textbf{PP} & 0.6/-0.2 & 57.5/13.2 & 9.4/7.4 & 17.7/-0.5& 7.1/6.5 & 1.5/-0.2 & 45.1/13.7 & 15.8/18.6 & 12.1/3.2 & 12.2/16.8 \\
            \bottomrule
        \end{tabular}}        
    \end{subtable}
\end{table}
\subsubsection{Fragility of high-capacity detectors to geometric perturbations.}
The sensitivity of both legacy (CP) and contemporary (FF) architectures to subtle geometric coordinate shifts, induced by PB and NE attacks, substantiates our hypothesis that high-capacity detectors over-rely on canonical spatial distributions and fine-grained local features to achieve peak performance. This dependency creates a critical representational bottleneck, whereby models prioritize the memorization of specific PC representations over the learning of robust volumetric invariants. This vulnerability is particularly acute within safety-critical classes, where inherent semantic sparsity and structural rigidity make detectors highly sensitive to coordinate shifts. 

\subsubsection{Sensitivity of pillar-based detectors to feature injection.} The pillar-based models exhibit strong robustness against perturbation attacks, but the coarser spatial quantization in pillar encoding makes them susceptible to point injections. The models demonstrate high sensitivity against PA attacks and while the failure of PP is anticipated due to its simplified encoder, the parallel degradation of the more sophisticated PN architecture suggests that the underlying pillar-based abstraction, inherited from its predecessor—constitutes a fundamental and shared structural vulnerability. The poor performance on safety-critical classes is attributed to the low geometric complexity of these objects. The pedestrian and cyclist classes occupy few pillars after encoding and have distinct vertical signatures. Post point-injection, this latent representation is disturbed, resulting in higher misclassifications. On the bright side, we observe that although the PN model uses CenterHead as \emph{detection head}, which demonstrated poor resilience against NE attack, PN exhibits strong robustness against NE attack. This underlines that pillar encoding combined with tailored architectural choices, which in case of PN is pre-trained ConvNeXt \emph{backbone}, effectively addresses the bottlenecks of high-capacity detection frameworks.

\subsubsection{PC density as a double-edge sword.} Contrary to the intuition that a higher PC density might bolster model resilience, our findings establish that density and robustness are largely orthogonal. While the sparsity of nuScenes, with a $32$-beam LiDAR, makes it fragile to PD attack due to the loss of essential geometric landmarks, the rich spatial structure of Waymo, with a $64$-beam LiDAR, fails to mitigate perturbation attacks (PB, NE). In fact, we argue that a denser PC offers a richer geometric landscape for adversarial manipulation without significantly altering its global occupancy. The empirical results confirm that, while increased sensor resolution improves performance on clean datasets, the resulting dense PCs are paradoxically more susceptible to fine-grained structural transformations.
\begin{figure}[tb]
  \centering
  \includegraphics[height=5.15cm]{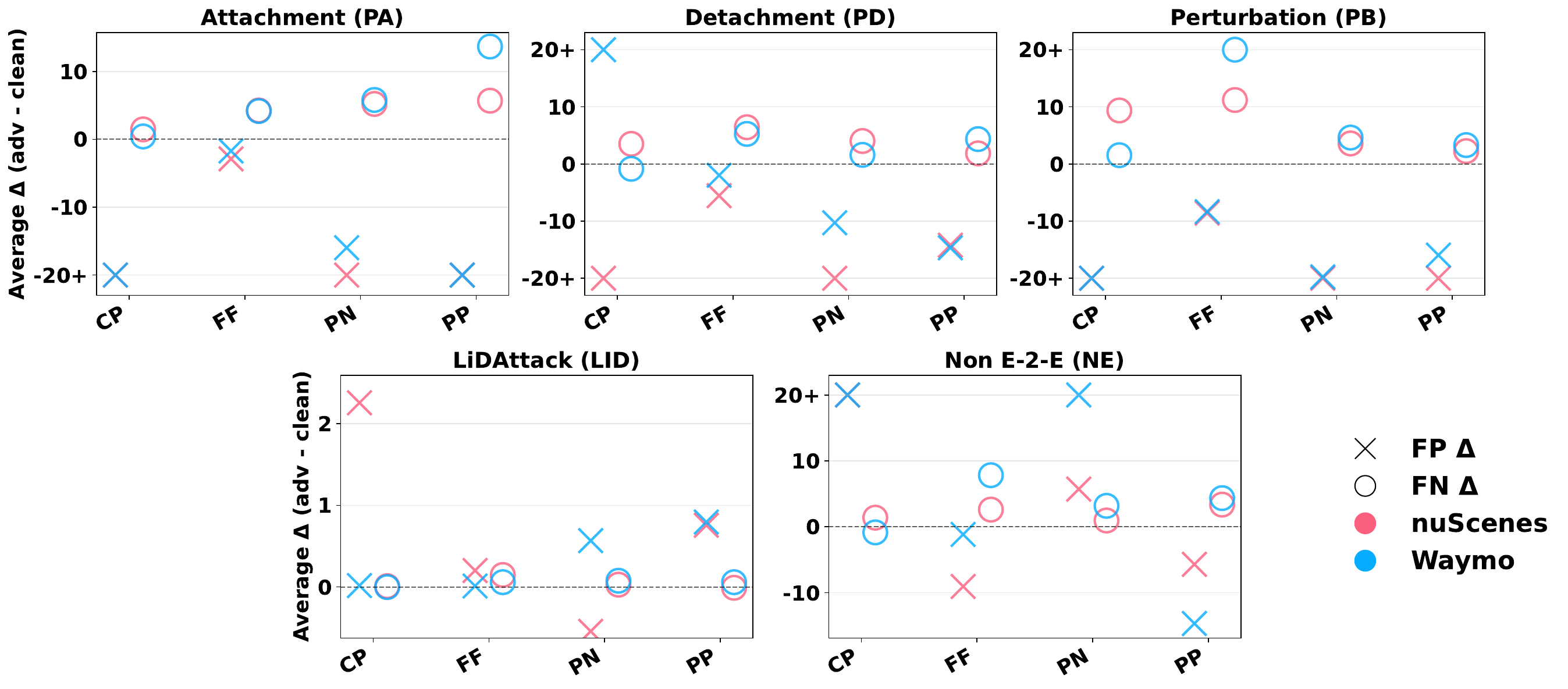}
  \caption{Visualization of average change ($\triangle$) in false positives (FP) and false negatives (FN) for each model on nuScenes and Waymo datasets.}
  \label{fig:4_impact_classification}
\end{figure}
\subsubsection{Strong robustness against black-box attack}
Across all models and datasets, the LiD attack remains the least effective, with performance drops typically staying below 2.6\%. This shows that the models are robust to black-box attacks and cannot be easily fooled by a universal noise pattern generated in such attacks.

\subsubsection{Adversarial Erasure and Systemic Blindness.} 
We quantify the shift in FPs and FNs ($\triangle$) in \cref{fig:4_impact_classification}. To establish a robust baseline, we first compute the mean FP and FN counts across the benign (clean) data distribution for each architecture. Subsequently, we evaluate these metrics under adversarial conditions to
\begin{wrapfigure}[16]{r}{0.27\textwidth}
  \centering
  \includegraphics[width=0.25\textwidth]{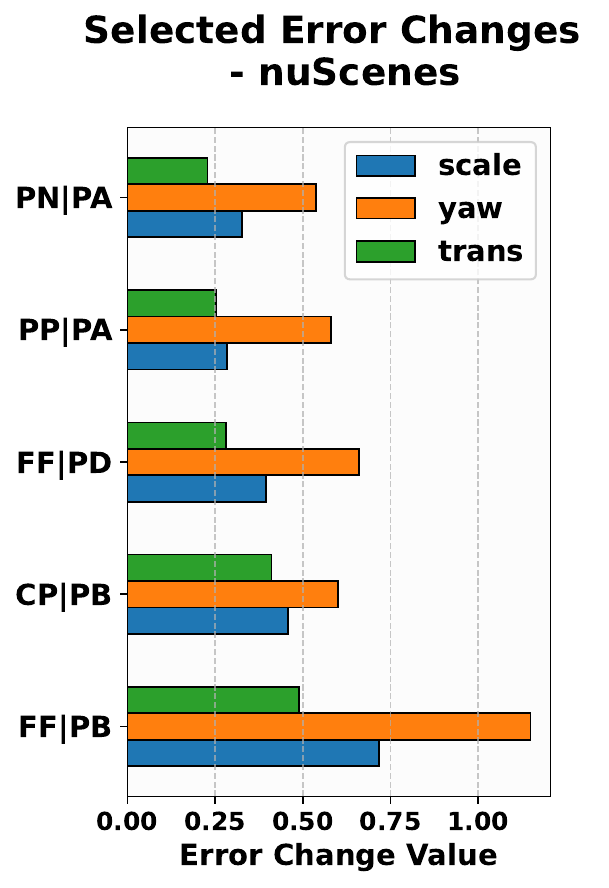}
  \caption{Comparison of geometric errors for each model|attack combination on nuScenes.}
  \label{fig:4_localization_err}
\end{wrapfigure}
determine the directional bias of the induced errors. Our results indicate that these attacks are predominantly erasure-dominant, resulting in a state of systemic blindness. The transformer-based FF designed to mitigate FN in high capacity detectors, in contrast, generates high FNs in PB attack. Although FF leverages long-range attention to refine predictions and identify \emph{hard instances}, its global attention introduces a structural vulnerability. The model's expansive receptive field allows adversarial noise from distant, non-target points to propagate into the feature representations of legitimate objects. Consequently, the same mechanism intended to improve model's recall facilitates the transmission of noise, effectively compromising the model's robustness. \\
\textbf{Discussion.} The heatmap-based (CP) or hard instance-based (FF) detection approach used by high capacity detectors shows strong limitations in identifying perturbed representations. Contrary to the performance on clean dataset, pillar-based models are more robust than voxel-based high-capacity detectors demonstrating that the vertical aggregation inherent in pillar encoding provides a degree of structural invariance that more granular voxel features lack. In general, advanced 3D-ODs suffer from systemic blindness when the underlying PC structure is transformed and are particularly susceptible to objects possessing distinct vertical representations, such as \emph{safety critical} classes.

\subsection{Impact on Localization}

Beyond categorical accuracy, to investigate the geometric fidelity of true positives, we quantify the localization error induced by adversarial perturbations on correctly predicted objects. This error is decomposed into three fundamental spatial components: volumetric scaling (scale), angular orientation (yaw) and spatial translation (trans). As visualized in \cref{fig:4_localization_err}, our analysis on nuScenes reveals that yaw estimation is disproportionately sensitive to adversarial noise, which disrupts orientation reasoning even when the semantic identity of an object is preserved. Notably, we observe an architectural divergence in robustness: while FF under PD attack yields fewer mispredictions, the resulting bounding boxes exhibit profound misorientation. In contrast, CP and FF, despite their extreme mAP sensitivity to NE attacks, maintain higher localization precision on the true positive detections. This confirms that mAP-centric evaluation for adversarial robustness is insufficient and an efficient benchmark must include the impact on localization errors.\\
\textbf{Discussion.} The evaluation of impact on localization errors exposes how modern 3D-OD's vulnerabilities lead to spatial misorientation in predicted boxes. From a system perspective, these errors are critical; downstream motion planners rely on precise orientation to infer an object's intent and predicted trajectory. Consequently, a failure of 3D-OD propagates through the stack, potentially leading to risky planning decisions.

\subsection{Impact on Distance}
\begin{figure}[tb]
    \centering
 
    % First graph
    \begin{subfigure}{\textwidth}
        \centering
        \includegraphics[width=0.99\linewidth]{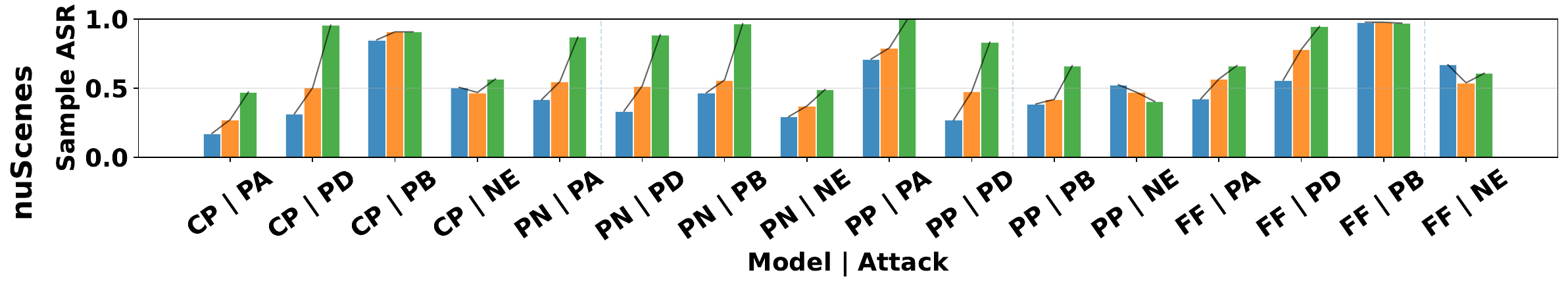}
        %\caption{nuScenes}
        \label{fig:adv_noise}
    \end{subfigure}
    
    \begin{subfigure}{\textwidth}
        \centering
        \includegraphics[width=0.99\linewidth]{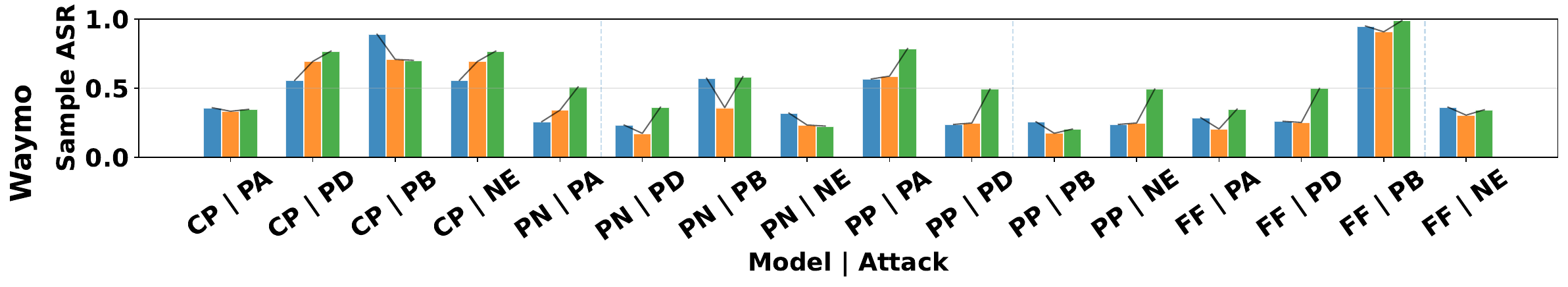}
        %\caption{Waymo}
        \label{fig:pillar_feat}
    \end{subfigure}   
    \caption{Impact on ASR for near-far objects for each model-attack pair. The proximity of the objects is represented as \protect\legsquare{nearblue} : Near, \protect\legsquare{midorange} : Mid and \protect\legsquare{fargreen} : Far.}
    \label{fig:4_impact_distance}
\end{figure}
Since the PC density has inverse correlation to the proximity of object from ego, it is hypothesized that objects in the immediate vicinity, characterized by higher sampling resolution, should demonstrate superior resilience to adversarial noise. To empirically validate this hypothesis, we categorize the objects into three bins - near, mid and far and then evaluate the impact on their prediction. Referring to the official nuScenes documentation for detection range, we classify objects within 20m from the ego as \emph{near}, 20-35m as \emph{mid} and objects within 35-50m as \emph{far} for both nuScenes and Waymo dataset. The results of our study, as visualized in \cref{fig:4_impact_distance}. Although the initial hypothesis holds in most cases, for perturbation attacks (PB and NE), the adversarial vulnerabilities are distance-agnostic. These findings underscore the limitations of mAP-centric evaluation methods to measure adversarial robustness.\\ 
\textbf{Discussion.}
Our empirical analysis refutes the intuitive assumption, that Near-ego objects are imminently less vulnerable. The attacks expose vulnerabilities in recent 3D-OD which are sensitive to subtle shifts in geometric representations irrespective of their distance from ego. The global attention and iterative refinement (HIP) of FF makes it equally vulnerable to attacks on near-ego objects, similar to far-away objects. Based on these findings, we strongly underscore the necessity for distance-aware learning strategies \cite{distance_aware_1, distance_aware_2, distance_aware_3} that reinforce both the high-density near-field and the sparse far-away representations.

\subsection{Impact of PC Density}
\begin{figure}[tb]
  \centering
  \includegraphics[height=3cm]{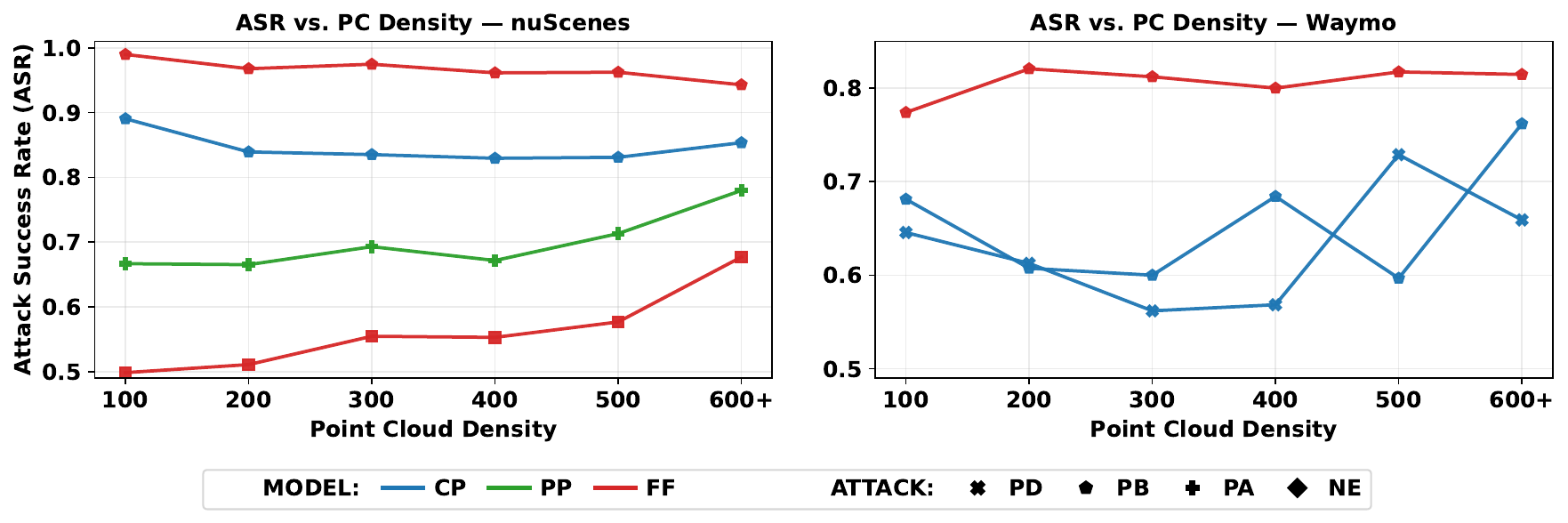}
  \caption{Visualization of ASR vs point cloud density of models under different attacks.}
  \label{fig:4_impact_pc_density}
\end{figure}
To validate our hypothesis that a \emph{holistic robustness evaluation requires the bifurcation of structural precision from predictive confidence}, we analyze performance degradation subject to PC density of adversarial samples in \cref{fig:4_impact_pc_density}. In this 
\begin{wrapfigure}[15]{r}{0.28\textwidth}
  \centering
  \includegraphics[width=0.28\textwidth]{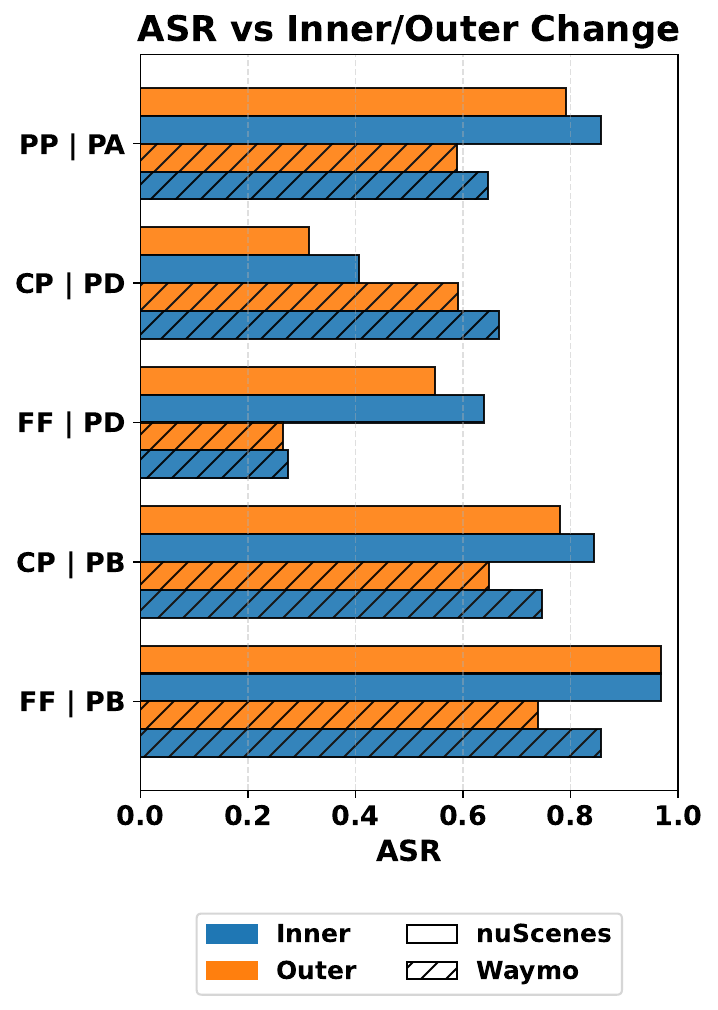}
  \caption{Impact of PC localization on ASR.}
  \label{fig:4_pc_inn_out}
\end{wrapfigure}
context, density is defined as the cardinal point count within an object’s (\eg \emph{Car}) volumetric bounds following adversarial manipulation. We observe that few model-attack pairs (FF-PB, CP-PB and PP-PA) report density-invariant susceptibility, analogous to observations in \cref{tab:4_imp_class}. We observe a paradoxical inverse correlation between PC density and adversarial vulnerability in specific model-attack combinations (PP-NE and CP-NE for nuScenes and CP-PA and PN-PB for Waymo), which substantiates that structural robustness is an inherent property of architectural design rather than a simple byproduct of input sparsity. For most other combinations, the robustness improves as the PC becomes denser (check Appendix). Although this sounds intuitive, we argue that \emph{mAP} alone cannot provide these insights, therefore validating our hypothesis.

\subsection{Impact of PC localization}
\label{sec:4_results_pc_loc}

To further evaluate structural robustness, we analyze the impact on ASR by partitioning the PCs into concentric spatial regions \emph{inner} and \emph{outer}. Our hypothesis posits that perturbations along the points in the inner box strongly degrade performance as they provide volumetric density to the structural contours. The empirical results in \cref{fig:4_pc_inn_out}, representing top-5 most affected combinations, validate our hypothesis. In most cases, ASR is similarly high for \emph{inner} and \emph{outer} points, indicating perturbations along edges and surface result in similar degradation. However, when the points are dropped along the surface, PD attack, induces significantly higher degradation compared to dropping peripheral points. These results once again validate our previous claim - evaluating structural robustness is equally important for a holistic evaluation of adversarial robustness.

\subsection{Sensitivity Analysis}
\label{sec:4_sensitivity_analysis}

\subsubsection{Scaling threshold for Impact of Point Cloud localization.} We parameterize the inner-outer point cloud split using a concentric scaling threshold applied to the ground truth bounding box. To ensure a balanced representation across 
\begin{wrapfigure}[16]{r}{0.35\textwidth} 
  \centering
  \begin{subfigure}{\linewidth}
    \centering
    \includegraphics[width=0.99\linewidth]{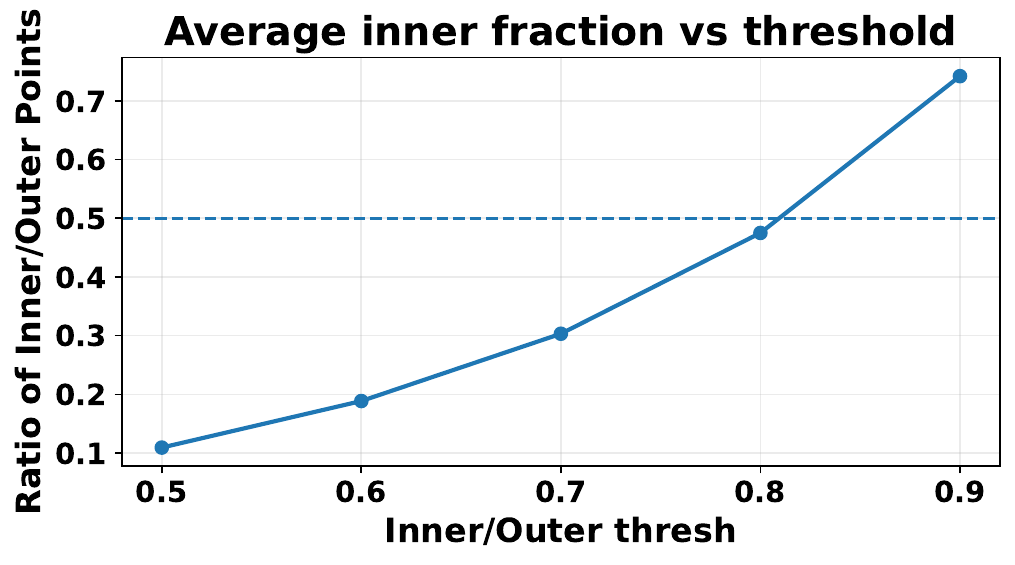}
    \caption{nuScenes}
    \label{fig:emp_inn_out_nusc}
  \end{subfigure}

  \begin{subfigure}{\linewidth}
    \centering
    \includegraphics[width=0.95\linewidth]{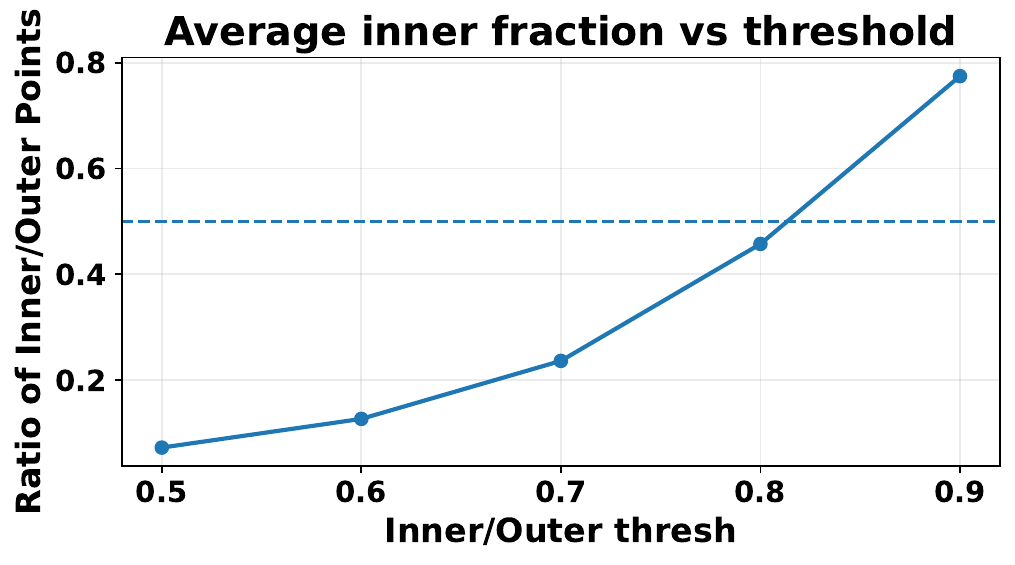}
    \caption{Waymo}
    \label{fig:emp_inn_out_way}
  \end{subfigure}
  
  \caption{Mean of ratio between inner and outer point clouds across all objects for nuScenes and Waymo dataset.}
  \label{fig:app_emp_inn_out}
\end{wrapfigure}
varying levels of LiDAR sparsity, we calibrate the threshold to equi-partition the points between the internal core and the peripheral volume.  Empirical results in \cref{fig:app_emp_inn_out} across both datasets indicate that a threshold of $0.8$ successfully yields a near-equal point distribution, mitigating density-driven bias in our localized robustness evaluation.

\begin{table}[t]
    \centering
    \caption{Sensitivity analysis of \emph{Confidence} threshold.}
    \label{tab:side_by_side_metrics}
    
    % --- Sub-table 1: Avg. ASR vs Threshold ---
    \begin{subtable}{0.44\textwidth}
        \centering
        \caption{Avg. ASR over all models at various \emph{Confidence} thresholds.}
        \label{tab:side_by_side_metrics_a}
        \setlength{\tabcolsep}{2.5pt}
        \scriptsize % Native, crisp font sizing without horizontal stretching
        \begin{tabular}{l c >{\columncolor{gray!20}}c c c c}
            \toprule
            \textbf{Attack} & \multicolumn{5}{c}{\textbf{Confidence Threshold}} \\
            \cmidrule(lr){2-6}
             & \textbf{.10} & \textbf{.15} & \textbf{.17} & \textbf{.20} & \textbf{.25} \\
            \midrule
            \textbf{NE} & .481 & .481 & .493 & .512 & .543 \\
            \textbf{PA} & .498 & .498 & .511 & .530 & .562 \\
            \textbf{PB} & .702 & .702 & .711 & .724 & .742 \\
            \textbf{PD} & .488 & .488 & .499 & .515 & .541 \\
            \bottomrule
        \end{tabular}       
    \end{subtable}
    \hfill % Pushes the subtables neatly to the margins
    % --- Sub-table 2: Model Thresholds ---
    \begin{subtable}{0.54\textwidth}
        \centering
        \caption{Absolute difference b/w standard ASR and our ASR on Waymo.}
        \label{tab:side_by_side_metrics_b}
        \setlength{\tabcolsep}{2pt}
        \scriptsize % Native, crisp font sizing without horizontal stretching
        \begin{tabular}{l cccc cccc}
            \toprule
             & \multicolumn{4}{c}{\textbf{\textsf{Threshold = 0.15}}} & \multicolumn{4}{c}{\textbf{\textsf{Threshold = 0.20}}} \\
            \cmidrule(lr){2-5} \cmidrule(lr){6-9}
            \textbf{Model} & \textbf{NE} & \textbf{PA} & \textbf{PB} & \textbf{PD} & \textbf{NE} & \textbf{PA} & \textbf{PB} & \textbf{PD} \\
            \midrule
            \textbf{CP} & .01 & .00 & .01 & .01 & .03 & .01 & .04 & .07 \\
            \textbf{FF} & .01 & .00 & .02 & .00 & .04 & .01 & .05 & .03 \\
            \textbf{PN} & .00 & .02 & .00 & .00 & .01 & .07 & .00 & .00 \\
            \textbf{PP} & .00 & .01 & .00 & .00 & .01 & .04 & .00 & .00 \\
            \bottomrule
        \end{tabular}       
    \end{subtable}
\end{table}

\subsubsection{Selection of \emph{Confidence} threshold for ASR calculation.} To justify the choice of the 0.15 confidence threshold within our ASR metric, we perform a comprehensive sensitivity analysis across a spectrum of thresholds from 0.10 to 0.25 as shown in \cref{tab:side_by_side_metrics}. Setting an excessively high threshold, such as 0.17 or 0.20, introduces an overly pessimistic bias that misclassifies functional detections as total failures. In practical autonomous driving architectures, downstream planning systems typically ingest proposals down to a 0.10 or 0.15 cutoff to maintain high recall. If a higher threshold is enforced during evaluation, any adversarial perturbation causing a minor, non-critical confidence drop (e.g., to 0.18) is incorrectly penalized as a complete detection failure, artificially inflating the ASR. As demonstrated in \cref{tab:side_by_side_metrics_a}, the average ASR remains completely flat between 0.10 and 0.15, validating 0.15 as the optimal empirical boundary that tightly couples adversarial success with genuine operational risk. This is further supported by \cref{tab:side_by_side_metrics_b}, where the absolute difference between our metric and the standard ASR at the 0.15 threshold is negligible ($\le 0.02$) across all models.

%% file: sec/5_conclusion.tex
\section{Conclusion}
\label{sec:conclusion}

In this study, we conduct a comprehensive adversarial robustness nalysis of SoTA 3D-OD models across commonly benchmarked datasets using latest adversarial attacks. Our analysis, which holistically evaluates the impact of adversarial attacks on various structural and predictive factors, exposes the limitations of existing \emph{mAP-dominant} frameworks to measure adversarial robustness. The key finding from our empirical results is that the path toward safe autonomous perception requires a shift beyond maximizing \emph{mAP} on clean data and prioritizing structurally robust \emph{encoder} and \emph{head} architectures. Our study reveals that, although recent models significantly outperform classical models on clean data, they lack the ability to detect subtle volumetric inconsistencies in object representations. The empirical results on predictive robustness validate our first hypothesis that SoTA 3D-OD architectures are fragile against coordinated point cloud perturbations. Furthermore, on the dataset-level, our investigation highlights that adversarial robustness is orthogonal to the point cloud density as the models trained on Waymo, providing richer spatial features, experience high sensitivity to point cloud transformations. Finally, our experiments on distance-specific robustness reveal that when subjected to point cloud perturbation attacks, models exhibit low robustness even for near-ego objects. Based on our extensive study, we propose the following recommendations to the research community: 1) Incorporating \emph{distance-aware} learning strategies during model training. 2) To compensate for the geometric homogeneity in point cloud representations, future models should incorporate adversarial data augmentation techniques in their training. We hope that our analysis provides a strong foundation for holistically evaluating adversarial robustness in LiDAR-based 3D object detectors.

\subsubsection{Limitations.} We trained some models from scratch, where implementation from original authors was missing, under commonly used training configurations. Although we assume that the results will not be fundamentally different, we acknowledge that the training settings might not be optimal, and there is certainly room for improvement. 

\section*{Acknowledgements}
The project on which this publication is based was co-funded by the European Union under grant number EFRE-20801104. We are also grateful to the support provided by Intelligent Sensing, Diagnostics and Prognostics Research Lab ISDPRL and the Digital Research Alliance of Canada AllianceCan. We thank Mitacs for supporting Mr. Yerdana Maulenbay through the RISE-Globalink Research Internship program [IT45510]. Additionally, we thank the RISE (Research Internships in Science and Engineering) Germany program organized by DAAD (Deutscher Akademischer Austauschdienst). Finally, we would like to thank Dr. Ling Bai for their review of the paper, and for offering constructive criticism which helped to address shortcomings in this paper. The computations at the University of Wuppertal were carried out on the PLEIADES cluster, which was supported by the Deutsche Forschungsgemeinschaft (DFG, grant No. INST 218/78-1 FUGG) and the Bundesministerium für Bildung und Forschung (BMBF).